\documentclass[letterpaper, 10 pt, conference]{ieeeconf}  
\IEEEoverridecommandlockouts                              
\usepackage{graphicx} 
\usepackage{svg}
\usepackage{booktabs}

\usepackage{amsmath} 
\usepackage{amssymb}  
\usepackage{babel}
\usepackage{url}

\title{\LARGE \bf
 StixelNExT: Toward Monocular Low-Weight Perception for Object Segmentation and Free Space Detection
}

\author{Marcel Vosshans$^{1,2}$, Omar Ait-Aider$^{2}$, Youcef Mezouar$^{2}$ and Markus Enzweiler$^{1}$
\thanks{$^{1}$The authors are with the Institute for Intelligent Systems which is part of the Faculty of Computer Science and Engineering, University of Applied Sciences Esslingen, Germany
        {\tt\small \{marcel.vosshans, markus.enzweiler\}@hs-esslingen.de}}%
\thanks{$^{2}$The authors are with the Institut Pascal ISPR (Image, Systems of Perception, Robotics), Universite Clermont Auvergne INP / CNRS, France
        {\tt\small \{youcef.mezouar, omar.ait-aider\}@uca.fr}}%
}

\begin{document}
\maketitle
\thispagestyle{empty}
\pagestyle{empty}

\begin{abstract}
In this work, we present a novel approach for general object segmentation from a monocular image, eliminating the need for manually labeled training data and enabling rapid, straightforward training and adaptation with minimal data. 
Our model initially learns from LiDAR during the training process, which is subsequently removed from the system, allowing it to function solely on monocular imagery. 
This study leverages the concept of the Stixel-World to recognize a medium level representation of its surroundings. 
Our network directly predicts a 2D multi-layer Stixel-World and is capable of recognizing and locating multiple, superimposed objects within an image. 
Due to the scarcity of comparable works, we have divided the capabilities into modules and present a free space detection in our experiments section. 
Furthermore, we introduce an improved method for generating Stixels from LiDAR data, which we use as ground truth for our network.
\end{abstract}

\section{INTRODUCTION}
Self-driving vehicles, heralded as the future of transportation, rely on intricate software stacks for their autonomous operation. 
As articulated by Mobileye \cite{mobileye_vision_technologies_ltd_how_2023} and other industry leaders, these stacks encompass various stages, with perception being one of the earliest and foundational components. 
In the realm of perception, a critical juncture arises where the quality of information significantly influences the overall performance of autonomous vehicles, due to error propagation. 
In robotics and intelligent vehicles, there is a constant trade-off between performance and runtime, determined by computational power. 
Medium-level representations play a crucial role in this trade-off, especially in Advanced Driver-Assistance Systems (ADAS), where one key concept is the Stixel-World \cite{cordts_stixel_2017}. 
This method primarily utilizes stereo vision, enabled by dual camera systems, to selectively extract only the essential data from the camera for understanding traffic scenes. 
It strikes a balance between detail capture and computational efficiency. 
In our work, we propose an approach for 2D multi-layer Stixel prediction, solely based on monocular images.
This contrasts with LiDAR-based methods, as vision-based data offers a denser array of information, more amenable to semantic classification, providing a distinct advantage over the sparser data typically derived from LiDAR. 
In its current state, our work's output representation focuses on multi-layer object segmentation and does not recover depth estimates. 
The segmentation results are obtained without any manual annotation, solely based on data derived from LiDAR.

\begin{figure}
    \centering
    \includegraphics[width=\linewidth]{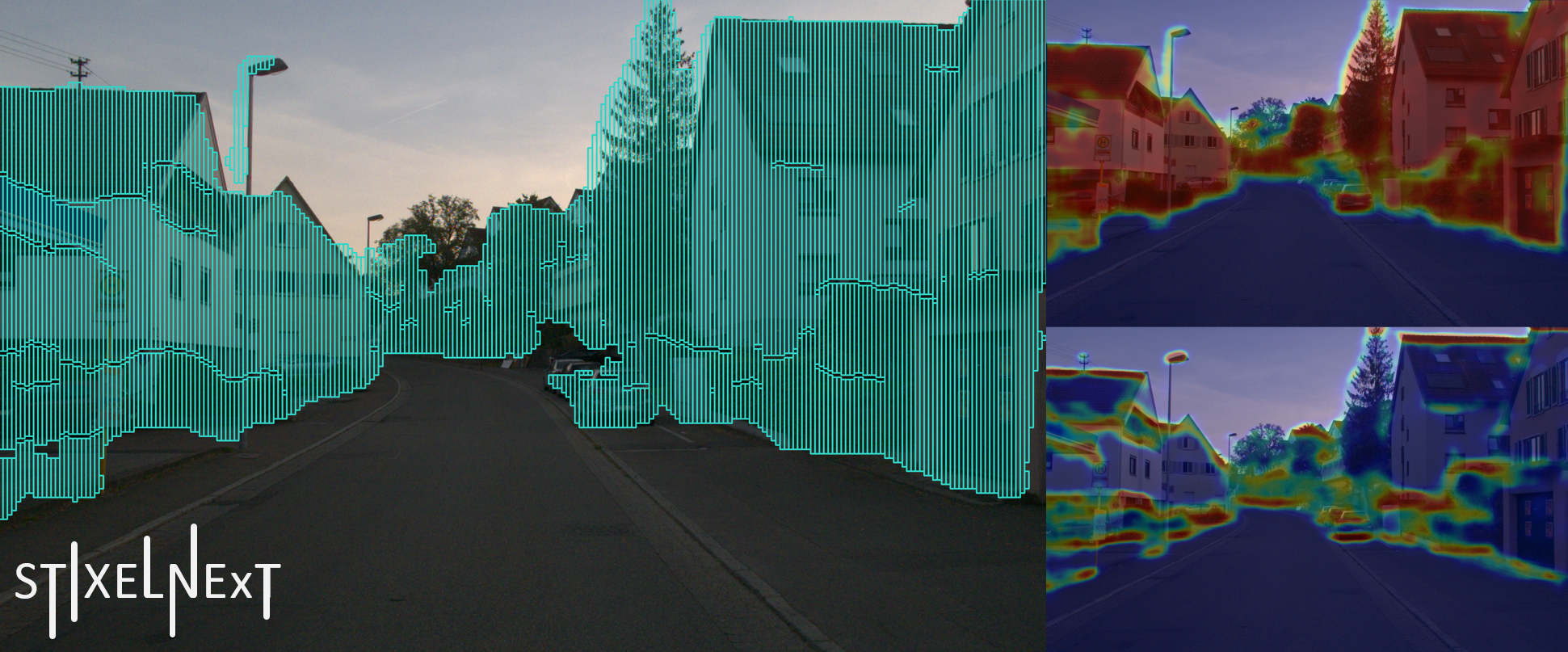}
    \caption{A \textbf{multi-layer Stixel result of StixelNExT}, along with the corresponding heat maps (right, upper: object occupancy, lower: object differentiation). The image on the left shows both the passable area and the segmented objects.}
    \label{fig:multi-layer Stixel}
\end{figure}

Inspired by StixelNet \cite{levi_stixelnet_2015}, which demarcates passable areas from obstacles using a demarcation line across the image, our research extends this concept by shifting the focus towards segmentation. 
This enables the detection of unknown objects, and allows  the determination of their height in pixel, column-wise, akin to Stixels, as depicted in Fig. \ref{fig:multi-layer Stixel}. 
This advancement marks a progression from StixelNet \cite{levi_stixelnet_2015}, evolving towards a more nuanced and detailed perception through the determination of multiple Stixels per column. 
In this work, we aim to introduce the segmentation aspect of our Stixel-World generation approach, which utilizes a single monocular image.
Another cornerstone of our research is the generation of ground truth data.
While StixelNet utilized LiDAR-derived ground truth, our method significantly enhances this process to accommodate the complexities introduced by the concept of multi-layer Stixels.
This improvement is crucial because the quality and accuracy of training data are paramount in artificial intelligence, especially in specialized domains that differ from conventional challenges like object detection or semantic segmentation. 
Given the cost and labor intensiveness of manual labeling, the exploration of fast automated labeling techniques becomes necessary.
It addresses a critical bottleneck in the field, automatic labeling research has the potential to significantly expedite and reduce the cost of developing ADAS technologies.

Generalization remains a persistent challenge in AI, often resulting in reduced performance, especially in computationally constrained environments.
This challenge is particularly critical in the context of ADAS, where rapid and accurate perception is essential. 
With an automatic labeling and learning pipeline, our system offers an efficient solution for adapting to new domains, making it a valuable asset in addressing generalization issues across various environments.
Our holistic perception strategy is well-suited to the dynamic and increasingly complex landscape of autonomous driving. 
On one hand, as complexity continues to grow, machine learning approaches become essential for handling it effectively.
On the other hand, traditional class-based solutions require us to anticipate and cover all imaginable cases to ensure accurate decision-making. 
Building upon these considerations, our goal is to present a comprehensive and versatile solution that effectively addresses these challenges, offering a general working approach.

\subsection{Contributions}
Our key contributions in this paper are as follows:
\begin{itemize}
    \item neural network-based prediction of multi-layer Stixels from a monocular camera
    \item novel approach for automated LiDAR-based ground truth generation for object segmentation
\end{itemize}
\subsection{Public Code Repositories}
The following public code repositories are available:
\begin{itemize}
    \item \textbf{Automatic Ground Truth Generator}\footnote{\url{https://github.com/MarcelVSHNS/StixelGENerator}}
    \item \textbf{StixelNExT}\footnote{\url{https://github.com/MarcelVSHNS/StixelNExT}} + Dataset (KITTI)
    \item \textbf{Evaluation Metric}\footnote{\url{https://github.com/MarcelVSHNS/StixelNExT-Eval}} 
\end{itemize}
\subsection{Paper Organisation}
In this paper, we present our development process of StixelNExT in a chronological sequence. 
We begin with ground truth generation, a critical influence on the network's performance.
We then delve into the network's architecture and mathematical foundations. 
Next, we evaluate its performance on our proprietary stereo camera dataset and compare it against an independent dataset with manually annotated labels to benchmark it against other neural networks.

\section{RELATED WORK}
The Stixel representation, originally defined as a compact medium-level representation of 3D environmental information \cite{denzler_stixel_2009}, has played a fundamental role in image processing and, notably, in vehicle automation. 
This form of representation has evolved from its basic format, which has enabled efficient environmental modeling, through several evolutionary steps. 
The extension to multi-layer Stixels \cite{pfeiffer_towards_2011} allowed for a more detailed capture of complex scenes, while the integration of semantic \cite{schneider_semantic_2016} (instance semantic \cite{hehn_instance_2019}) labels increased the density of information. 
With the introduction of Slanted Stixels \cite{hernandez-juarez_slanted_2019}, an even more realistic geometric adaptation was achieved.

A notable extension in Stixel research is the development of the Mono-Stixel \cite{brickwedde_mono-stixels_2019} approach. 
This approach sets itself apart by utilizing neural networks, diverging from traditional methodologies. 
Unlike conventional methods, the Mono-Stixel approach generates a Stixel world using multiple neural networks. 
The input for these networks is a combination of an RGB image, optical flow data, and semantic segmentation, which necessitates extensive preliminary processing. 
Despite the complexity of the input data, the network successfully aggregates depth, color, and semantic information to create a Stixel world. 
In a follow-up paper \cite{brickwedde_exploiting_2019}, the authors of the Mono-Stixel approach additionally integrated single-image depth prediction.
This enhancement allows for environmental perception in static scenes, in contrast to classic Structure-from-Motion approaches.

The work of StixelNet \cite{levi_stixelnet_2015} presented a novel network architecture initially based on the Visual Geometry Group (VGG) \cite{simonyan_very_2014} model. 
The architecture was fundamental in predicting Stixels using monocular camera input.
A key achievement of StixelNet was its capability to predict the end of freespace, a task comparable to that of stereo Stixel vision systems.
Notably, StixelNet employed a gradient method, informed by LiDAR data, to generate these contact points.
This marked an advancement in automated ground truth (AGT) data generation for training purposes. 
In their subsequent work \cite{garnett_real-time_2017} they refined this approach by integrating a general obstacle detector with a class-based object detector.  
Additionally, by incorporating a Single Shot Detector (SSD) + Pose into their network, they further expanded the system's ability to interpret scenes, still using a single RGB input. 
Among other improvements, a critical finding was that the performance enhancements were more due to the improved quality of ground truth data, rather than substantial changes to the network architecture.
Importantly, they demonstrated the system's real-time capabilities, achieving performance at 30 frames per second. 
StixelNet's approach, however, remains limited to 2D data and does not provide depth information.
Considering the distinct focus of StixelNet on road segmentation and Mono-Stixel on depth estimation via Structure-from-Motion, there are few comparable works in the field.
This fact is further underscored by our endeavor to align Stixel adaptation more closely with segmentation, highlighting the importance of integrating these two aspects for enhanced performance. 

In our experiments section, we evaluate a free space detection. It should be noted that there are significantly more networks available for Free-Space Detection \cite{yao_estimating_2015, fan_learning_2022}, some of which are also publicly accessible, including self-supervised models \cite{sanberg_free-space_2016}.

\section{GROUND TRUTH GENERATION}
The advantages of automated ground truth are evident: meticulous setup and the development of an algorithmic system can yield an almost inexhaustible dataset for training purposes. 
Particularly for tasks that diverge from conventional methods, such as generating a Stixel-World, access to ground truth data is limited, often necessitating reliance on simulations \cite{hernandez-juarez_gpu-accelerated_2017} with virtual Stixel generators. 
Our decision to employ automated ground truth (AGT) derived from real-world data is driven by the objective to achieve the capability of domain adaptation. 
An example illustrating the use of AGT is StixelNet. 
StixelNet generates Stixels using a gradient-based method following for each given column \(x\) as the derivative \( \frac{\delta D(x,y)}{\delta y}\).
This approach works fairly reliably for determining the initial contact point of an object up to a certain range. 
However, challenges arise with more distant and multiple objects per column: minor inclines in LiDAR measuring points lead to a lower gradient so its harder to differentiate between ground and objects to extract the Stixel as ground truth. 
Additional complexities occur in scenarios with terrain changes, like slopes or hills, as these also weaken the gradient, thus impeding the precise identification of contact points \cite{piewak_improved_2018}.
Our solution involves approximating and decomposing the overall problem into sub-problems, which enables us to address each issue individually. 
\begin{figure}
    \centering
    \includegraphics[width=0.96\linewidth]{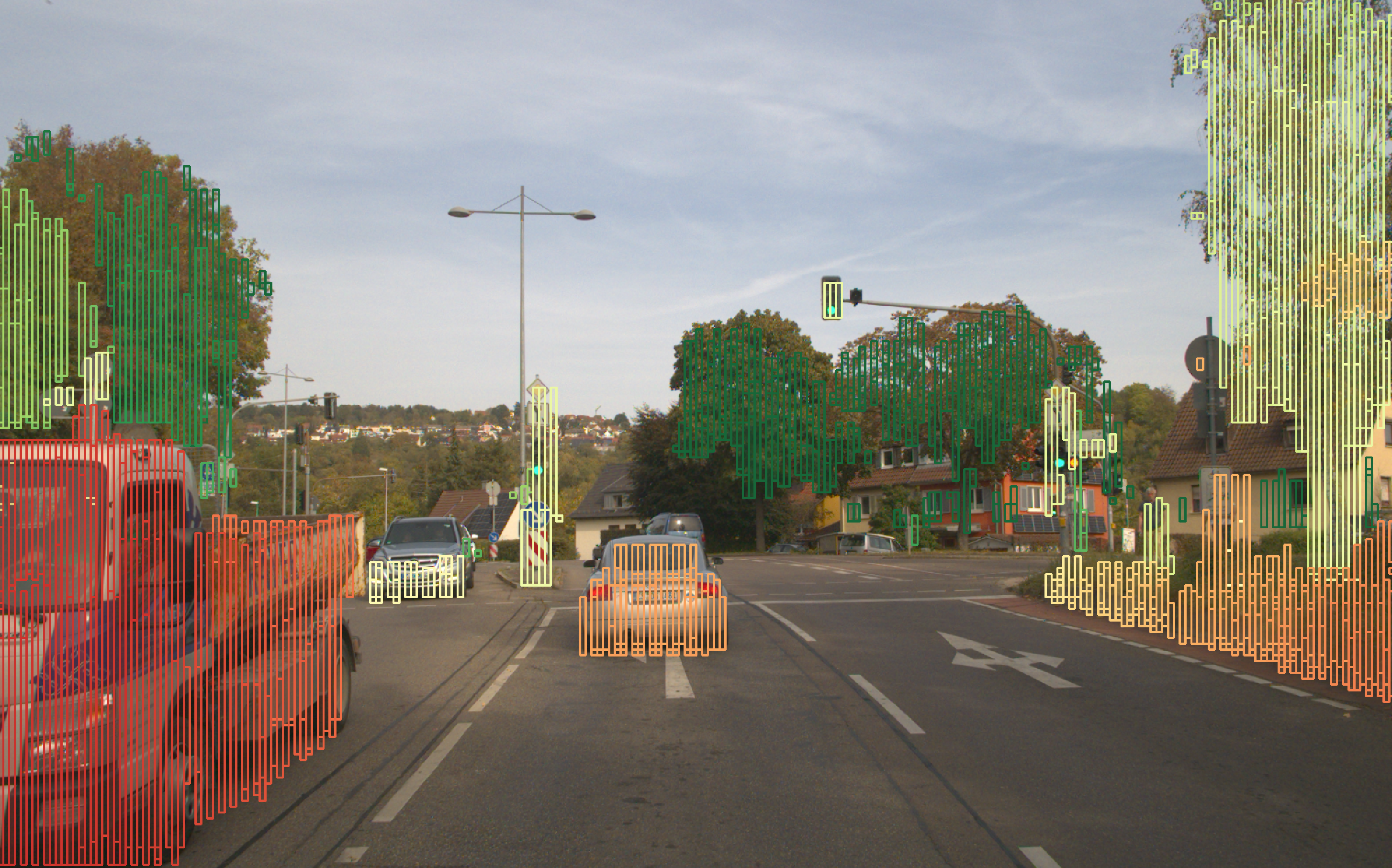}
    \caption{In the \textbf{AGT Stixel-World result}, only ground and swib (somewhere in between) objects are displayed. The color, ranging from red to green, encodes the distance from the sensor origin to the top point of a Stixel.}
    \label{fig:GT_stixel}
\end{figure}

In practical terms, this involves initially identifying the ground using a two-stage RANSAC-algorithm \cite{fischler_random_1987}, treating it as a linear basis.
Due to the differences in perspectives between the two sensors used, namely LiDAR and camera, and their extrinsic properties, there exists a variation in the viewpoints.
To address this, we have eliminated points (by the method of \cite{katz_direct_2007}) in the image which are obscured by objects over which the LiDAR can see, but the camera cannot.
This process necessitates intrinsic and extrinsic calibration information. Subsequently, we employed clustering techniques like the DBSCAN algorithm \cite{ester_density-based_1996} to divide the LiDAR point cloud based on angular resolution.
We parameterize the DBSCAN and RANSAC algorithms based on a qualitative evaluation over multiple samples.
Each angle, in turn, is clustered again according to its elevation into objects.
Depending on the cluster's distance from the origin, objects are categorized into ground objects and swib (somewhere in between) objects. 
Moving forward in the analysis, the highest point of an object and its curvature in the direction of ascent becomes the top point of a Stixel, standing on the previously assumed linear plane equation.
Swib objects, on the other hand, have their base point at the optical line of sight relative to the preceding object.
This is accomplished through the coordinate transformation of LiDAR points onto the RGB image.
\begin{table}[t]
\centering
\caption{Definition of Stixel types with their geometry and assigned semantic class for the free space detection evaluation}
\begin{tabular}{p{0.18\linewidth}cp{0.5\linewidth}}
Stixel type & Geometry & Semantic class following \cite{cordts_cityscapes_2016} \\
\midrule
ground \(\mathbb{G}\) & lying & ground, road, sidewalk, parking, rail track, terrain \\
ground objects \(\mathbb{GO}\), swib objects \(\mathbb{SO}\) & upright  & building, wall, fence, guard rail, bridge, tunnel, pole, pole group, traffic light, traffic sign, vegetation, person, rider, car, truck, bus, caravan, trailer, train, motorcycle, bicycle \\
sky \(\mathbb{S}\) & end of data & sky \\
\label{tab:semantics}
\end{tabular}
\end{table}
Through this approach, we were able to completely describe each image in terms of Stixels based on LiDAR-camera projection, as demonstrated in Table \ref{tab:semantics}.
Some of the results are showcased in Figure \ref{fig:GT_stixel}. 
Through the consistent use of approximation methods, our approach is robust against non-planar road geometries, takes extrinsic calibration into account, and is fully scalable.

For this study, we generated our own dataset, which is intended to be published in full extent.
Our setup primarily consisted of a custom-built stereo camera (for evaluation against the conventional Stixel algorithm), using the left camera as input, an Ouster OS-1 with 128 layers for LiDAR ground truth data generation, and an Inertial Navigation System (INS) for correcting self-motion.
The sensors were fully calibrated and synchronized, both spatially and temporally.
To gather data for this dataset, we drove for approximately one hour through various scenarios, including urban, countryside, and highway environments.

\section{StixelNExT}
Our presented StixelNExT model aims to predict a multi-layer Stixel world, albeit in 2D.
Therefore, our approach has been to maintain the essential characteristics of information reduction found in stereo-based Stixel generation, viewing it as a selective down sampler and extractor of desired data from a 3D environment.

\subsection{Architecture}
\begin{figure}
    \centering
    \includegraphics[width=0.96\linewidth]{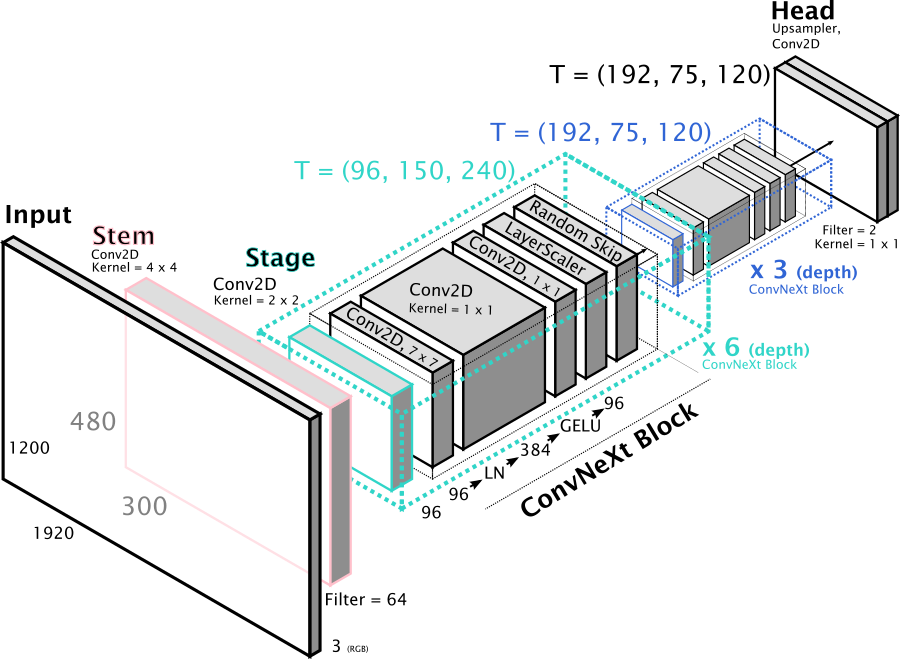}
    \caption{The \textbf{StixelNExT architecture}, as a distillation of ConvNeXt \cite{liu_convnet_2022}, is depicted in this schematic diagram. The model, consisting of 1.5 million parameters, showcases its scalability by illustrating the replication of blocks as the scale increases.}
    \label{fig:architecture}
\end{figure}

Our StixelNExT model is inspired by and partially adapted from the ConvNeXt \cite{liu_convnet_2022} architecture.
Additionally, variations of the backbone were explored, including derivatives with ResNet \cite{he_deep_2016} and Inception \cite{szegedy_going_2015}. 
However, these also resulted in similar, marginal differences as noted by StixelNet \cite{garnett_real-time_2017}.
We have designed the output to be a multiple of the original format - the resolution -, often by a factor of \(\frac{1}{4}\) (KITTI dataset \cite{geiger_vision_2013}) or \(\frac{1}{8}\) (our dataset), allowing it to overlay onto the original image like a grid.
The primary objectives of the network are two-fold: firstly, to represent an occupancy grid indicating the presence or absence of objects in the category object \(\mathbb{GO}\) and \(\mathbb{SO}\) (see Table \ref{tab:semantics}), and secondly, to identify the outer horizontal edges (top and bottom points) of each Stixel with the object differentiation matrix.
Hence, we have two channels in the output layer of our architecture.
This process of two-dimensional object segmentation followed by splitting the Stixel into differentiated object Stixel is done by post-processing.
We have distilled the ConvNeXt model to its essentials, adopting only the initial portions in accordance with the convolution factors, and shaping it with a suitable head (see Fig. \ref{fig:architecture}).
In our studies, we investigated two models: one with \(0.27\) million parameters and another with \(1.5\) million parameters (for comparison, StixelNet had about \(30\) million parameters and was capable of operating at 30 FPS). 
The difference arises from the number of depth layers and the implemented stages.
For simplicity, we present the larger network here, although the smaller one has also achieved similar results. 
For our unpublished dataset, this results in an input resolution of \(h = 1200\) and \(w = 1920\), leading to an output tensor of \(T = (2, 150, 240)\) at a resolution of \(8\) px width per Stixel.
For our KITTI adaptation, the input dimensions are \(h = 376\) and \(w = 1248\) and the output respectively \(T = (2, 94, 312)\) at a resolution of \(4\) px.
For the interpretation of the Stixels, we normalized the heat maps to set a uniform threshold.
\begin{figure*}[ht]
    \centering
    \includegraphics[width=\textwidth]{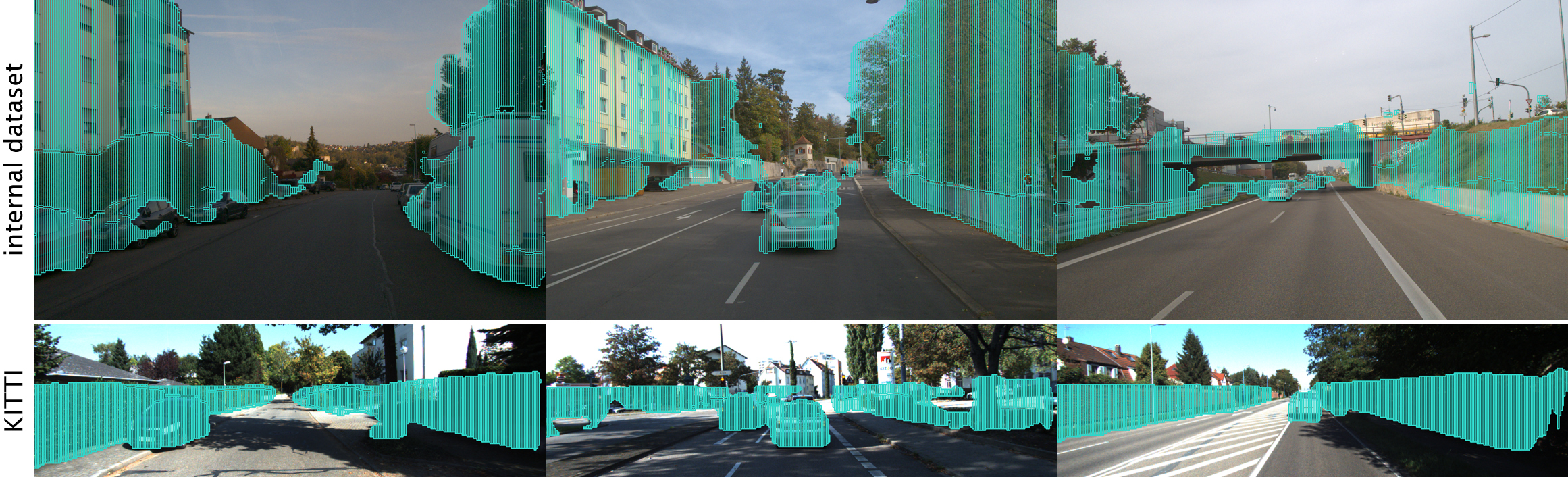}
    \caption{\textbf{Multi-layer Stixel results from StixelNExT.} The upper row displays results trained with our internal dataset, while the lower row showcases results trained on the KITTI dataset. A unique feature of our dataset is the capture of LiDAR points far above the horizon, allowing the camera to effectively learn from the LiDAR. In the case of KITTI, it's evident that the network's output is constrained by the limited vertical field-of-view of the KITTI LiDAR.}
    \label{fig:results}
\end{figure*}
Figure \ref{fig:results} presents a selection of results and Figure \ref{fig:multi-layer Stixel} illustrates a sample along with the corresponding heat maps.
Through hyperparameter tuning, we identified two stages as the ideal trade-off, with the number of depth layers being six and three respectively.
The stem features were kept relatively small at \(64\).
Notably, we train the network from scratch each time, without leveraging any pre-trained weights.

\subsection{Loss Function}
When developing the loss function for StixelNExT, we initially used established functions as a guide before starting to fine-tune it.  
For the occurrence matrices \(T\), we employed a separate Binary Cross Entropy loss, commonly known as \cite{paszke_pytorch_2019}:
\begin{align}
L_{BCE}(y, \hat{y}) &= -\frac{1}{N} \sum_{i=1}^{N} L_i \label{eq:total_loss} \\
L_i &= y_i \log(\hat{y}_i) + (1 - y_i) \log(1 - \hat{y}_i) \notag
\end{align}
While \(y_i\) represents the target and \(\hat{y}_i\) the prediction,  each is represented as element \(T_{uv}\) with \(u\) as columns and \(v\) as rows. 
Additionally, with the goal of creating distinct thresholds, we attempted to reduce the total number of predictions to bolster the model's confidence, particularly for object transitions.
This was achieved using a simple summation technique, which minimizes the total number of predictions and leads to clearer confidences:
\begin{equation}
L_{Sum}(y, \hat{y}) = -\frac{1}{N} \sum_{u=1}^{m} \sum_{v=1}^{n} T_{uv}
\end{equation}
With respect to the output tensor \(T\), we refer to the first dimension (see Fig. \ref{fig:multi-layer Stixel}) as occupancy \(M_{occ}\) and as cut (object differentiation) \(M_{cut}\) matrix. 
The occupancy matrix determines the presence of an object based on confidence levels, while the object differentiation identifies object borders.
Consequently, the following overall loss formula was established:
\begin{equation}
L(y, \hat{y}) = \alpha \cdot L_{BCE_{occ}} + \beta \cdot L_{Sum_{occ}} + \gamma \cdot L_{BCE_{cut}}
\end{equation} 
Particularly for predictions representing positions in an image, an immediate evaluation is required.
Without a defined threshold, it becomes challenging to determine what should be counted as relevant.
This means that, to reflect a priori knowledge such as continual shapes of objects in the loss function, a threshold for a Stixel position must be specified, thereby eliminating any mathematical derivability. 
Resorting to assessing probability brings us back to the Binary Cross Entropy (BCE) Loss.
A direct end-to-end prediction might have circumvented this challenge; however, we will explore this aspect in more detail in our experiments section.

\subsection{Baseline and Datasets}
For our baseline, we utilized a publicly available implementation of the multi-layer Stixel paper \cite{gishi523_multi-stixel-world_2019}, making several adjustments to align it with our approach.
This baseline involves the generation of disparity images and the creation of a Stixel-world using dynamic programming, with full consideration of the extrinsic setup. 
For the implementation of StixelNet, we used a publicly available implementation \cite{tran_obstacle_2020} as well and made necessary adaptations, including the addition of data loaders for the datasets and configurations. 
StixelNet operates at a resolution of \((370, 800)\).

The data foundation for this work was somewhat limited due to this chosen baseline, as a dataset incorporating both LiDAR and stereo camera was necessary.
Publicly, this includes datasets like KITTI \cite{geiger_vision_2013}, but primarily, we conducted evaluations on our own custom created dataset.
Regarding the split of the KITTI dataset, we followed StixelNet's suggestion for the validation dataset. Additionally, we used the following sequences (09\_26\_d\_79, 09\_26\_d\_96, 09\_28\_d\_38, 09\_30\_d\_16) as test datasets for evaluation. 
Both networks are trained and evaluated on their corresponding datasets.

\section{EXPERIMENTS}
In this section, we delve into a series of experiments conducted to thoroughly investigate and assess various aspects of our model.

\subsection{The Naive Evaluation}
\begin{table}[b]
\centering
\caption{Evaluation on the test split of the internal dataset (2040 samples) and the KITTI dataset (964 samples) \cite{geiger_vision_2013}\\ with \(IoU = 0.5\) and varying threshold \(t\)}
\begin{tabular}{lccc}
 & Precision & Recall & F1 \\
\multicolumn{4}{l}{internal dataset} \\
\midrule
StixelNExT, \(t=0.32\) & 0.2931 & \textbf{0.2391} & 0.2546 \\
StixelNExT, \(t=0.45\) & 0.3348 & 0.2291 & \textbf{0.2639} \\
StixelNExT, \(t=0.68\) & \textbf{0.3796} & 0.1948 & 0.2507 \\
Stereo & 0.1431 & 0.2325  & 0.1691  \\
& & & \\
\multicolumn{4}{l}{KITTI dataset} \\
\midrule
StixelNExT, \(t=0.42\) & 0.5347 & \textbf{0.1651} & 0.2505 \\
StixelNExT, \(t=0.58\) & 0.5589 & 0.1644 & \textbf{0.2522} \\
StixelNExT, \(t=0.70\) & \textbf{0.5854} & 0.1559 & 0.2447 \\
Stereo & 0.1265 & 0.0857  & 0.1011  \\
\label{tab:self-evaluation}
\end{tabular}
\end{table}
As mentioned earlier, our aim is to compare our results with multi-layer Stixel representations, for which there is no directly comparable work available to date.
For evaluation, we opted for an approach similar to object detection, assessing the predicted precision of the model against the actual objects using an Intersection over Union (IoU) metric.
From an evaluation strategy perspective, a Stixel can be considered as an one-dimensional object, equal to a bounding box.
This approach allows us to draw parallels with established metrics in object detection, while adapting them to the unique characteristics of Stixel-based representations. 
The resulting metric led to a varying precision-recall curve, from which we derived the ideal operating point. 
We then pitted this derived operating point against the baseline using the same IoU methodology.
In Table \ref{tab:self-evaluation}, we present some selected numeric results of our evaluation.
A few notes on the evaluation process: it maps the set of ground truth Stixels to predictions and attempts to find the best matching Stixel in the column, subsequently checking for an \(IoU >= 0.5 \). 
Compared to the qualitative results (see Fig. \ref{fig:results}), the precision-recall metric does not reflect the performance of the model accurately. However, it helps in identifying the optimal operating point. 
The lower numerical results can also be attributed to noise in the ground truth data, particularly with vegetation. 
In such cases, numerous Stixels may be identified, but they might not always have precise relevance in terms of segmentation. 
This contrasts with scenarios involving vehicle Stixels, where an object of interest is detected but represented with only a few Stixels. 
Furthermore, it's worth mentioning that the precision of the stereo algorithm is subject to a negative offset due to noise in disparity, often producing effects similar to those in the vegetation of ground truth data.
This phenomenon is particularly noticeable in areas like the sky, adding an additional layer of complexity to the evaluation process.
This is the reason why we refer to our baseline as the "naive evaluation." In fact, there is no directly comparable system for this application.
\begin{figure}
    \centering
    \includegraphics[width=\linewidth]{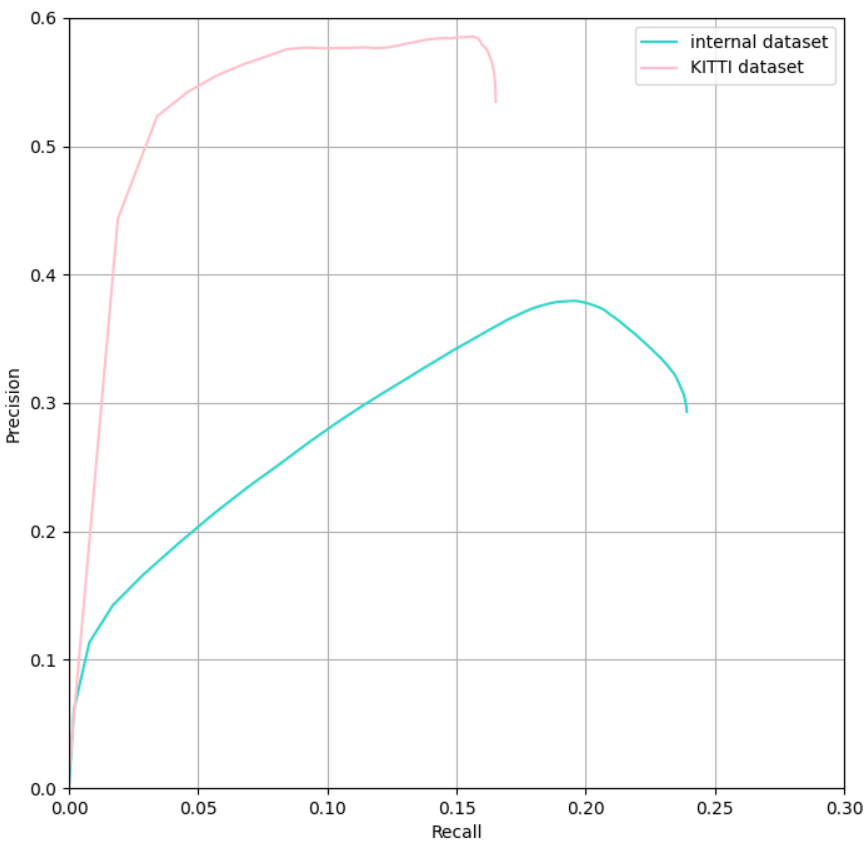}
    \caption{The two Graphs illustrate the \textbf{progression of the precision-recall} curve with varying thresholds. Turquoise shows the results on our internal dataset and pink the model trained on the KITTI dataset. The thresholds \(t\) start right and increase to the left.}
    \label{fig:prec-recall-curve}
\end{figure}
In Fig. \ref{fig:prec-recall-curve}, it is evident that the progression of the precision-recall curve is atypical.
One explanation is the abstraction of the Stixel into heat maps which need to be interpreted. 
The varying threshold \(t\), defined as the numerical border for detecting a Stixel from the occupancy matrix at a certain position, does not fundamentally alter the position of the Stixel in our case. 
Instead, it primarily influences the edge areas or the division into sub-Stixels. 
The threshold allows fine tuning of the desired results. At a low threshold value, broad Stixels and numerous splits are expected (both precision and recall are low).
As the threshold increases, the splits become fewer, and the model approaches its optimal operating point (F1 score). 
Beyond a certain point, no more Stixels are predicted, or splits are made (precision and recall decreases).
Consequently, its not possible to expect a typical progression of precision-recall here, since there is no variation in the Stixel predictions and concepts such as NMS (Non-Maximum Suppression) are not necessary/possible here. 
The Stereo Baseline has a fixed operating point due to the absence of a deployable threshold. It can be characterized by a precision of \(0.1265\) and a recall of \(0.0857\).
For further analysis, we chose the best F1 score (see Table \ref{tab:self-evaluation}) as operating point, aiming to balance precision and recall for optimal performance with \(t_{intern} = 0.45\) and \(t_{KITTI} = 0.58\).

\subsection{The Fairer Evaluation}
\begin{figure*}
    \centering
    \includegraphics[width=\textwidth]{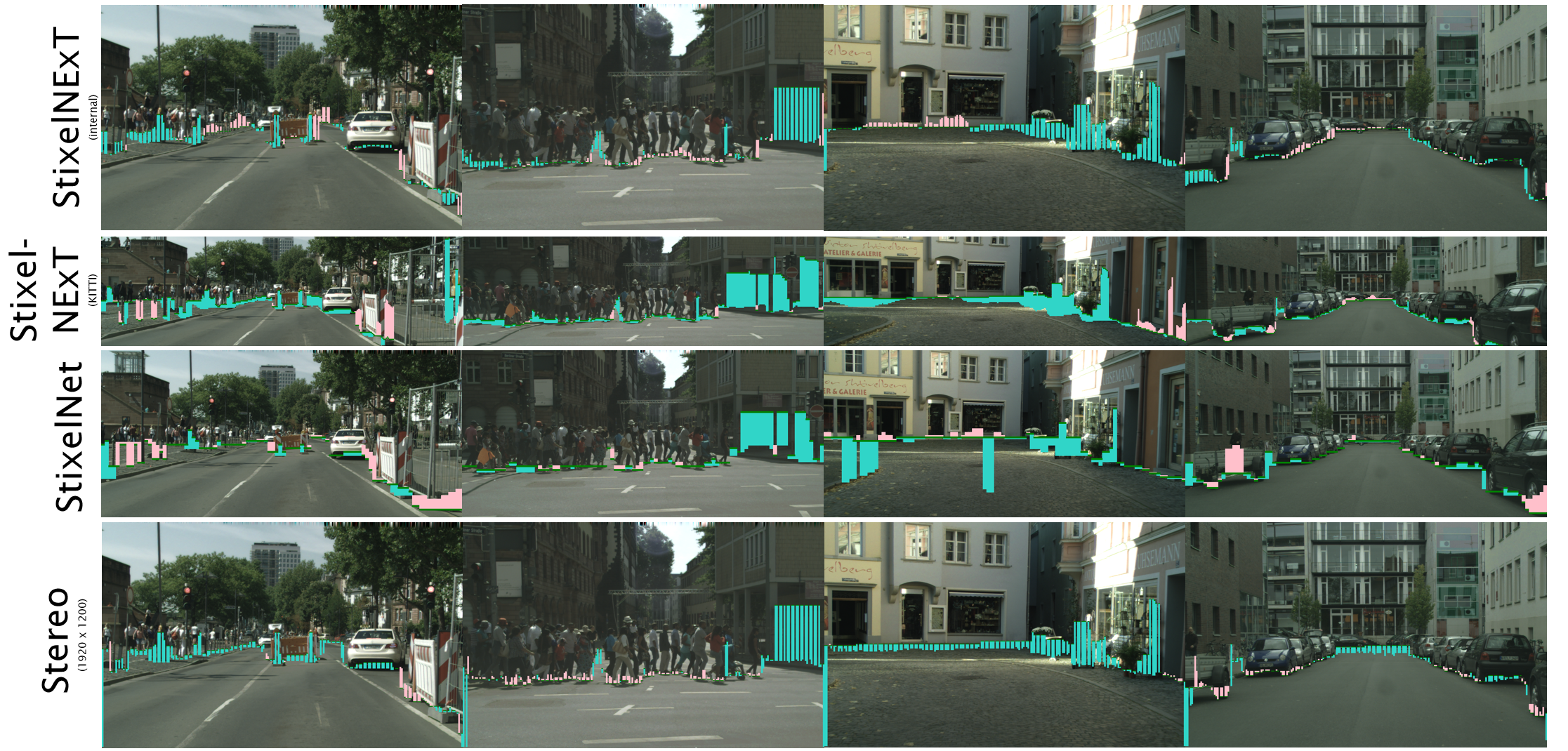}
    \caption{Some \textbf{results from the obstacle evaluation}, showcasing how different systems perform on the Cityscapes dataset in detecting open areas as per Table \ref{tab:semantics} (ground, ground objects). Pink colors indicate a positional delayed detection (with the prediction appearing first from top to bottom, followed by the ground truth), while turquoise signifies the opposite, representing an earlier detection for e.g. from a safety-critical perspective the safer system.}
    \label{fig:obstacle-detection}
\end{figure*}

In our quest for a comparable application, we primarily focused on breaking down the problem and its capabilities and evaluate them independently. 
Apart from semantic segmentation, which demands a more dedicated focus, we turned our attention to free space detection and conceived a "fairer evaluation" approach. 
However, this form of evaluation only represents a portion of the capabilities of StixelNExT and is therefore intended primarily as a comparable bridge to the foundational project, StixelNet. For the model, this means that we continue to predict the complete 2D Stixel world, but only evaluate the bottom point per column. 
Our aim was to demonstrate the versatility and efficacy of StixelNExT in comparison to these established methods, while acknowledging that our approach encompasses a broader range of use cases. 
For the evaluation, we included a third dataset that provides pixel-precise segmentation and stereo data, such as the Cityscapes dataset \cite{cordts_cityscapes_2016}. 
The idea was to categorize semantic labels into two groups: passable areas and obstacles (you can find a full list in Table \ref{tab:semantics}). 
This categorization aligns with the use case of StixelNet and can also be represented by both the stereo camera and StixelNExT, through the detection of the bottom point in each column. 
Figure \ref{fig:obstacle-detection} illustrates the comparison and results of the different systems, showcasing how each method performs under the same external evaluation criteria.
Due to heavy down-sampling or up-sampling of the image input sizes in the models, we decided to crop the images.

For the obstacle evaluation, we chose to use the validation split, as the test dataset does not have publicly available labels due to the challenges involved. This encompasses 500 fine annotated images with dimensions of \(h = 1024\) px and \(w = 2048\) px. 
We initially removed the ego-vehicle from the images, as it would have required learning an offset, and then created a crop with an ideal region of interest (ROI) and minimal scaling per model and image. 
In this process, each column of the image accrues points (on a 1:1 basis for every pixel from the top edge of the image to the ground contact point of the obstacle). 
The difference between each prediction and the ground truth is penalized in pixels and subsequently represented as a percentage. 
This metric offers a nuanced understanding of how closely our predictions align with the actual obstacles, providing a concrete measure of the model's accuracy in real-world scenarios compared to existing systems.     
\begin{table}[ht]
\centering
\caption{Obstacle detection by segmentation on Cityscapes dataset \cite{cordts_cityscapes_2016}, trained on different datasets. \(\Sigma\) represents the summed percentage differences divided by the number of columns, whereas \(\sigma\) denotes the mean percentage standard deviation.}
\begin{tabular}{lcc}
 & Score \(\Sigma\) [\%] & \(\sigma\) [\%]\\
\midrule
StixelNExT @ \(t = 0.45\), internal & 91.070 & 8.023 \\
StixelNExT @ \(t = 0.58\), KITTI & 91.661 & 4.967 \\
StixelNet, KITTI & \textbf{94.370} & 3.566 \\
Stereo \((1200 x 1920)\) px & 91.054 & \textbf{2.825} \\
Stereo \((370 x 800)\) px & 88.528 & 9.664 \\
\label{tab:obstacle-results}
\end{tabular}
\end{table}
In Fig. \ref{fig:obstacle-detection} and Table \ref{tab:obstacle-results}, we present a comparison between StixelNExT and StixelNet, alongside the conventional stereo baseline.
However, the significance of this evaluation is subject to some doubt. 
As seen in Fig. \ref{fig:obstacle-detection}, the semantic segmentation can vary, and depending on the chosen classes, one network may outperform the other.
It is important to note that while obstacle detection is merely a sub task for both StixelNExT and the Stereo baseline, it represents the intersection with StixelNet.
Although StixelNet remains a viable option for obstacle detection, StixelNExT offers multi-layer Stixels to detect objects with a 2D height, enabling more complex perception tasks."

\section{CONCLUSIONS AND FUTURE WORK}
Our work with StixelNExT has successfully demonstrated the feasibility of 2D multi-layer Stixel localization in images, achieved through an efficient training process that leverages LiDAR data without the need for manually annotated labels. 
While the current iteration of StixelNExT does not include depth prediction, this is an area where we have already seen preliminary successes and continue to focus our effort on. 
Although incorporating a mono depth estimator \cite{godard_digging_2018} or a scene flow \cite{brickwedde_mono-sf_2019} network could provide a more comprehensive solution, our approach intentionally minimizes reliance on complex post-processing to maintain simplicity and effectiveness.
This project primarily aimed to create a solid foundation by carefully designing the data set, evaluation metrics and training procedures. 
Looking ahead, enhancing StixelNExT with monocular depth estimation presents an intriguing avenue for future research and could significantly bolster the network's capabilities.

\addtolength{\textheight}{-10.6cm}

\bibliographystyle{IEEEtran}
\bibliography{references}

\begin{thebibliography}{10}
\providecommand{\url}[1]{#1}
\csname url@samestyle\endcsname
\providecommand{\newblock}{\relax}
\providecommand{\bibinfo}[2]{#2}
\providecommand{\BIBentrySTDinterwordspacing}{\spaceskip=0pt\relax}
\providecommand{\BIBentryALTinterwordstretchfactor}{4}
\providecommand{\BIBentryALTinterwordspacing}{\spaceskip=\fontdimen2\font plus
\BIBentryALTinterwordstretchfactor\fontdimen3\font minus \fontdimen4\font\relax}
\providecommand{\BIBforeignlanguage}[2]{{%
\expandafter\ifx\csname l@#1\endcsname\relax
\typeout{** WARNING: IEEEtran.bst: No hyphenation pattern has been}%
\typeout{** loaded for the language `#1'. Using the pattern for}%
\typeout{** the default language instead.}%
\else
\language=\csname l@#1\endcsname
\fi
#2}}
\providecommand{\BIBdecl}{\relax}
\BIBdecl

\bibitem{mobileye_vision_technologies_ltd_how_2023}
\BIBentryALTinterwordspacing
M.~V.~T. Ltd, ``How {Autonomous} {Vehicles} {Work}: the {Self}-{Driving} {Stack},'' May 2023. [Online]. Available: \url{https://www.mobileye.com/blog/autonomous-vehicle-day-the-self-driving-stack/}
\BIBentrySTDinterwordspacing

\bibitem{cordts_stixel_2017}
\BIBentryALTinterwordspacing
M.~Cordts, T.~Rehfeld, L.~Schneider, D.~Pfeiffer, M.~Enzweiler, S.~Roth, M.~Pollefeys, and U.~Franke, ``\BIBforeignlanguage{en}{The {Stixel} {World}: {A} medium-level representation of traffic scenes},'' \emph{\BIBforeignlanguage{en}{Image and Vision Computing}}, vol.~68, pp. 40--52, Dec. 2017. [Online]. Available: \url{https://linkinghub.elsevier.com/retrieve/pii/S0262885617300331}
\BIBentrySTDinterwordspacing

\bibitem{levi_stixelnet_2015}
\BIBentryALTinterwordspacing
D.~Levi, N.~Garnett, and E.~Fetaya, ``\BIBforeignlanguage{en}{{StixelNet}: {A} {Deep} {Convolutional} {Network} for {Obstacle} {Detection} and {Road} {Segmentation}},'' in \emph{\BIBforeignlanguage{en}{British {Machine} {Vision} {Conference} 2015}}.\hskip 1em plus 0.5em minus 0.4em\relax Swansea: British Machine Vision Association, 2015, pp. 109.1--109.12. [Online]. Available: \url{http://www.bmva.org/bmvc/2015/papers/paper109/index.html}
\BIBentrySTDinterwordspacing

\bibitem{denzler_stixel_2009}
\BIBentryALTinterwordspacing
H.~Badino, U.~Franke, and D.~Pfeiffer, ``The {Stixel} {World} - {A} {Compact} {Medium} {Level} {Representation} of the {3D}-{World},'' in \emph{Pattern {Recognition}}, J.~Denzler, G.~Notni, and H.~Süße, Eds.\hskip 1em plus 0.5em minus 0.4em\relax Berlin, Heidelberg: Springer Berlin Heidelberg, 2009, vol. 5748, pp. 51--60. [Online]. Available: \url{http://link.springer.com/10.1007/978-3-642-03798-6\_6}
\BIBentrySTDinterwordspacing

\bibitem{pfeiffer_towards_2011}
\BIBentryALTinterwordspacing
D.~Pfeiffer and U.~Franke, ``\BIBforeignlanguage{en}{Towards a {Global} {Optimal} {Multi}-{Layer} {Stixel} {Representation} of {Dense} {3D} {Data}},'' in \emph{\BIBforeignlanguage{en}{British {Machine} {Vision} {Conference} 2011}}.\hskip 1em plus 0.5em minus 0.4em\relax Dundee: British Machine Vision Association, 2011, pp. 51.1--51.12. [Online]. Available: \url{http://www.bmva.org/bmvc/2011/proceedings/paper51/index.html}
\BIBentrySTDinterwordspacing

\bibitem{schneider_semantic_2016}
\BIBentryALTinterwordspacing
L.~Schneider, M.~Cordts, T.~Rehfeld, D.~Pfeiffer, M.~Enzweiler, U.~Franke, M.~Pollefeys, and S.~Roth, ``Semantic {Stixels}: {Depth} is not enough,'' in \emph{2016 {IEEE} {Intelligent} {Vehicles} {Symposium} ({IV})}.\hskip 1em plus 0.5em minus 0.4em\relax Gotenburg, Sweden: IEEE, Jun. 2016, pp. 110--117. [Online]. Available: \url{http://ieeexplore.ieee.org/document/7535373/}
\BIBentrySTDinterwordspacing

\bibitem{hehn_instance_2019}
\BIBentryALTinterwordspacing
T.~M. Hehn, J.~F.~P. Kooij, and D.~M. Gavrila, ``Instance {Stixels}: {Segmenting} and {Grouping} {Stixels} into {Objects},'' in \emph{2019 {IEEE} {Intelligent} {Vehicles} {Symposium} ({IV})}.\hskip 1em plus 0.5em minus 0.4em\relax Paris, France: IEEE, Jun. 2019, pp. 2542--2549. [Online]. Available: \url{https://ieeexplore.ieee.org/document/8814243/}
\BIBentrySTDinterwordspacing

\bibitem{hernandez-juarez_slanted_2019}
\BIBentryALTinterwordspacing
D.~Hernandez-Juarez, L.~Schneider, P.~Cebrian, A.~Espinosa, D.~Vazquez, A.~M. Lopez, U.~Franke, M.~Pollefeys, and J.~C. Moure, ``Slanted {Stixels}: {A} way to represent steep streets,'' \emph{International Journal of Computer Vision}, vol. 127, pp. 1643--1658, 2019, publisher: arXiv Version Number: 1. [Online]. Available: \url{https://arxiv.org/abs/1910.01466}
\BIBentrySTDinterwordspacing

\bibitem{brickwedde_mono-stixels_2019}
\BIBentryALTinterwordspacing
F.~Brickwedde, S.~Abraham, and R.~Mester, ``Mono-{Stixels}: {Monocular} depth reconstruction of dynamic street scenes,'' in \emph{2018 {IEEE} {International} {Conference} on {Robotics} and {Automation} ({ICRA})}, 2019, pp. 3369--3375. [Online]. Available: \url{https://arxiv.org/abs/1908.02635}
\BIBentrySTDinterwordspacing

\bibitem{brickwedde_exploiting_2019}
\BIBentryALTinterwordspacing
Brickwedde, Abraham, and Mester, ``Exploiting {Single} {Image} {Depth} {Prediction} for {Mono}-stixel {Estimation},'' in \emph{Computer {Vision} – {ECCV} 2018 {Workshops}}, 2019. [Online]. Available: \url{https://link.springer.com/10.1007/978-3-030-11009-3\_14}
\BIBentrySTDinterwordspacing

\bibitem{simonyan_very_2014}
\BIBentryALTinterwordspacing
K.~Simonyan and A.~Zisserman, ``Very {Deep} {Convolutional} {Networks} for {Large}-{Scale} {Image} {Recognition},'' 2014, publisher: arXiv Version Number: 6. [Online]. Available: \url{https://arxiv.org/abs/1409.1556}
\BIBentrySTDinterwordspacing

\bibitem{garnett_real-time_2017}
\BIBentryALTinterwordspacing
N.~Garnett, S.~Silberstein, S.~Oron, E.~Fetaya, U.~Verner, A.~Ayash, V.~Goldner, R.~Cohen, K.~Horn, and D.~Levi, ``Real-{Time} {Category}-{Based} and {General} {Obstacle} {Detection} for {Autonomous} {Driving},'' in \emph{2017 {IEEE} {International} {Conference} on {Computer} {Vision} {Workshop} ({ICCVW})}.\hskip 1em plus 0.5em minus 0.4em\relax Venice: IEEE, Oct. 2017, pp. 198--205. [Online]. Available: \url{http://ieeexplore.ieee.org/document/8265242/}
\BIBentrySTDinterwordspacing

\bibitem{yao_estimating_2015}
\BIBentryALTinterwordspacing
J.~Yao, S.~Ramalingam, Y.~Taguchi, Y.~Miki, and R.~Urtasun, ``Estimating {Drivable} {Collision}-{Free} {Space} from {Monocular} {Video},'' in \emph{2015 {IEEE} {Winter} {Conference} on {Applications} of {Computer} {Vision}}.\hskip 1em plus 0.5em minus 0.4em\relax Waikoloa, HI, USA: IEEE, Jan. 2015, pp. 420--427. [Online]. Available: \url{http://ieeexplore.ieee.org/document/7045916/}
\BIBentrySTDinterwordspacing

\bibitem{fan_learning_2022}
\BIBentryALTinterwordspacing
R.~Fan, H.~Wang, P.~Cai, J.~Wu, M.~J. Bocus, L.~Qiao, and M.~Liu, ``Learning {Collision}-{Free} {Space} {Detection} {From} {Stereo} {Images}: {Homography} {Matrix} {Brings} {Better} {Data} {Augmentation},'' \emph{IEEE/ASME Transactions on Mechatronics}, vol.~27, no.~1, pp. 225--233, Feb. 2022. [Online]. Available: \url{https://ieeexplore.ieee.org/document/9360504/}
\BIBentrySTDinterwordspacing

\bibitem{sanberg_free-space_2016}
\BIBentryALTinterwordspacing
W.~P. Sanberg, G.~Dubbelman, and P.~H.~N. de~With, ``Free-{Space} {Detection} with {Self}-{Supervised} and {Online} {Trained} {Fully} {Convolutional} {Networks},'' 2016. [Online]. Available: \url{https://arxiv.org/abs/1604.02316}
\BIBentrySTDinterwordspacing

\bibitem{hernandez-juarez_gpu-accelerated_2017}
\BIBentryALTinterwordspacing
D.~Hernandez-Juarez, A.~Espinosa, J.~C. Moure, D.~Vazquez, and A.~M. Lopez, ``{GPU}-{Accelerated} {Real}-{Time} {Stixel} {Computation},'' in \emph{2017 {IEEE} {Winter} {Conference} on {Applications} of {Computer} {Vision} ({WACV})}.\hskip 1em plus 0.5em minus 0.4em\relax Santa Rosa, CA, USA: IEEE, Mar. 2017, pp. 1054--1062. [Online]. Available: \url{http://ieeexplore.ieee.org/document/7926705/}
\BIBentrySTDinterwordspacing

\bibitem{piewak_improved_2018}
\BIBentryALTinterwordspacing
F.~Piewak, P.~Pinggera, M.~Enzweiler, D.~Pfeiffer, and M.~Zöllner, ``Improved {Semantic} {Stixels} via {Multimodal} {Sensor} {Fusion},'' in \emph{German {Conference} on {Pattern} {Recognition}}, 2018, pp. 447--458. [Online]. Available: \url{https://arxiv.org/abs/1809.08993}
\BIBentrySTDinterwordspacing

\bibitem{fischler_random_1987}
\BIBentryALTinterwordspacing
M.~A. Fischler and R.~C. Bolles, ``\BIBforeignlanguage{en}{Random {Sample} {Consensus}: {A} {Paradigm} for {Model} {Fitting} with {Applications} to {Image} {Analysis} and {Automated} {Cartography}},'' in \emph{\BIBforeignlanguage{en}{Readings in {Computer} {Vision}}}.\hskip 1em plus 0.5em minus 0.4em\relax Elsevier, 1987, pp. 726--740. [Online]. Available: \url{https://linkinghub.elsevier.com/retrieve/pii/B9780080515816500702}
\BIBentrySTDinterwordspacing

\bibitem{katz_direct_2007}
\BIBentryALTinterwordspacing
S.~Katz, A.~Tal, and R.~Basri, ``\BIBforeignlanguage{en}{Direct visibility of point sets},'' in \emph{\BIBforeignlanguage{en}{{ACM} {SIGGRAPH} 2007 papers}}.\hskip 1em plus 0.5em minus 0.4em\relax San Diego California: ACM, Jul. 2007, p.~24. [Online]. Available: \url{https://dl.acm.org/doi/10.1145/1275808.1276407}
\BIBentrySTDinterwordspacing

\bibitem{ester_density-based_1996}
\BIBentryALTinterwordspacing
M.~Ester, H.-P. Kriegel, J.~Sander, and X.~Xu, ``A density-based algorithm for discovering clusters in large spatial databases with noise,'' in \emph{kdd}, vol.~96, 1996, pp. 226--231. [Online]. Available: \url{https://www.dbs.ifi.lmu.de/Publikationen/Papers/KDD-96.final.frame.pdf}
\BIBentrySTDinterwordspacing

\bibitem{cordts_cityscapes_2016}
\BIBentryALTinterwordspacing
M.~Cordts, M.~Omran, S.~Ramos, T.~Rehfeld, M.~Enzweiler, R.~Benenson, U.~Franke, S.~Roth, and B.~Schiele, ``The {Cityscapes} {Dataset} for {Semantic} {Urban} {Scene} {Understanding},'' in \emph{2016 {IEEE} {Conference} on {Computer} {Vision} and {Pattern} {Recognition} ({CVPR})}.\hskip 1em plus 0.5em minus 0.4em\relax Las Vegas, NV, USA: IEEE, Jun. 2016, pp. 3213--3223. [Online]. Available: \url{http://ieeexplore.ieee.org/document/7780719/}
\BIBentrySTDinterwordspacing

\bibitem{liu_convnet_2022}
\BIBentryALTinterwordspacing
Z.~Liu, H.~Mao, C.-Y. Wu, C.~Feichtenhofer, T.~Darrell, and S.~Xie, ``A {ConvNet} for the 2020s,'' in \emph{Conference on {Computer} {Vision} and {Pattern} {Recognition} ({CVPR})}, 2022, publisher: arXiv Version Number: 2. [Online]. Available: \url{https://arxiv.org/abs/2201.03545}
\BIBentrySTDinterwordspacing

\bibitem{he_deep_2016}
\BIBentryALTinterwordspacing
K.~He, X.~Zhang, S.~Ren, and J.~Sun, ``Deep {Residual} {Learning} for {Image} {Recognition},'' \emph{IEEE conference on computer vision and pattern recognition (CVPR)}, pp. 770--778, 2016, publisher: arXiv Version Number: 1. [Online]. Available: \url{https://arxiv.org/abs/1512.03385}
\BIBentrySTDinterwordspacing

\bibitem{szegedy_going_2015}
\BIBentryALTinterwordspacing
C.~Szegedy, {Wei Liu}, {Yangqing Jia}, P.~Sermanet, S.~Reed, D.~Anguelov, D.~Erhan, V.~Vanhoucke, and A.~Rabinovich, ``Going deeper with convolutions,'' in \emph{2015 {IEEE} {Conference} on {Computer} {Vision} and {Pattern} {Recognition} ({CVPR})}.\hskip 1em plus 0.5em minus 0.4em\relax Boston, MA, USA: IEEE, Jun. 2015, pp. 1--9. [Online]. Available: \url{http://ieeexplore.ieee.org/document/7298594/}
\BIBentrySTDinterwordspacing

\bibitem{geiger_vision_2013}
\BIBentryALTinterwordspacing
A.~Geiger, P.~Lenz, C.~Stiller, and R.~Urtasun, ``\BIBforeignlanguage{en}{Vision meets robotics: {The} {KITTI} dataset},'' \emph{\BIBforeignlanguage{en}{The International Journal of Robotics Research}}, vol.~32, no.~11, pp. 1231--1237, Sep. 2013. [Online]. Available: \url{http://journals.sagepub.com/doi/10.1177/0278364913491297}
\BIBentrySTDinterwordspacing

\bibitem{paszke_pytorch_2019}
\BIBentryALTinterwordspacing
A.~Paszke, S.~Gross, F.~Massa, A.~Lerer, J.~Bradbury, G.~Chanan, T.~Killeen, Z.~Lin, N.~Gimelshein, L.~Antiga, A.~Desmaison, A.~Köpf, E.~Yang, Z.~DeVito, M.~Raison, A.~Tejani, S.~Chilamkurthy, B.~Steiner, L.~Fang, J.~Bai, and S.~Chintala, ``{PyTorch}: {An} {Imperative} {Style}, {High}-{Performance} {Deep} {Learning} {Library},'' \emph{Advances in neural information processing systems}, vol.~32, 2019, publisher: arXiv Version Number: 1. [Online]. Available: \url{https://arxiv.org/abs/1912.01703}
\BIBentrySTDinterwordspacing

\bibitem{gishi523_multi-stixel-world_2019}
\BIBentryALTinterwordspacing
gishi523, ``Multi-{Stixel}-{World},'' GitHub repository, 2019. [Online]. Available: \url{https://github.com/gishi523/multilayer-stixel-world}
\BIBentrySTDinterwordspacing

\bibitem{tran_obstacle_2020}
\BIBentryALTinterwordspacing
B.~Tran, ``Obstacle {Detection} {With} {StixelNet},'' GitHub repository, 2020. [Online]. Available: \url{https://github.com/xmba15/obstacle\_detection\_stixelnet}
\BIBentrySTDinterwordspacing

\bibitem{godard_digging_2018}
\BIBentryALTinterwordspacing
C.~Godard, O.~Mac~Aodha, M.~Firman, and G.~Brostow, ``Digging {Into} {Self}-{Supervised} {Monocular} {Depth} {Estimation},'' in \emph{Proceedings of the {IEEE}/{CVF} international conference on computer vision}, 2018, pp. 3828--3838. [Online]. Available: \url{https://arxiv.org/abs/1806.01260}
\BIBentrySTDinterwordspacing

\bibitem{brickwedde_mono-sf_2019}
\BIBentryALTinterwordspacing
F.~Brickwedde, S.~Abraham, and R.~Mester, ``Mono-{SF}: {Multi}-{View} {Geometry} {Meets} {Single}-{View} {Depth} for {Monocular} {Scene} {Flow} {Estimation} of {Dynamic} {Traffic} {Scenes},'' in \emph{{IEEE}/{CVF} {International} {Conference} on {Computer} {Vision}}, 2019, pp. 2780--2790. [Online]. Available: \url{https://arxiv.org/abs/1908.06316}
\BIBentrySTDinterwordspacing

\end{thebibliography}

\end{document}